\documentclass[10pt,twocolumn,letterpaper]{article}

\usepackage{cvpr}              

\usepackage{graphicx}
\usepackage{amsmath}
\usepackage{amssymb}
\usepackage{booktabs}

\usepackage{caption}
\usepackage{amsmath}

\newtheorem{proposition}{Proposition}
\usepackage{algorithm}
\usepackage{makecell}
\usepackage{algorithmic}

\usepackage{enumitem}   

\usepackage{chngcntr}
\usepackage{apptools}

\AtAppendix{\counterwithin{proposition}{section}}

\usepackage[accsupp]{axessibility}

\usepackage[pagebackref,breaklinks,colorlinks]{hyperref}

\usepackage[capitalize]{cleveref}
\crefname{section}{Sec.}{Secs.}
\Crefname{section}{Section}{Sections}
\Crefname{table}{Table}{Tables}
\crefname{table}{Tab.}{Tabs.}

\def\cvprPaperID{1980} 
\def\confName{CVPR}
\def\confYear{2022}

\begin{document}

\title{Prompt Distribution Learning}

\author{Yuning Lu\textsuperscript{1}\footnotemark[1], Jianzhuang Liu\textsuperscript{2}, Yonggang Zhang\textsuperscript{1}, Yajing Liu\textsuperscript{1}, Xinmei Tian\textsuperscript{1}\footnotemark[2]\\
\textsuperscript{1}University of Science and Technology of China\\
\textsuperscript{2}Huawei Noah's Ark Lab\\
{\tt\small \{lyn0,yonggang,lyj123\}@mail.ustc.edu.cn, liu.jianzhuang@huawei.com, xinmei@ustc.edu.cn}
}

\maketitle

\begin{abstract}

We present prompt distribution learning for effectively adapting a pre-trained vision-language model to address downstream recognition tasks.
Our method not only learns low-bias prompts from a few samples but also captures the distribution of diverse prompts to handle the varying visual representations.
In this way, we provide high-quality task-related content for facilitating recognition.
This prompt distribution learning is realized by an efficient approach that learns the output embeddings of prompts instead of the input embeddings.
Thus, we can employ a Gaussian distribution to model them effectively and derive a surrogate loss for efficient training.
Extensive experiments on 12 datasets demonstrate that our method consistently and significantly outperforms existing methods. For example, with 1 sample per category, it relatively improves the average result by $9.1$\% compared to human-crafted prompts.

\end{abstract}

\section{Introduction}
\label{sec:intro}

\renewcommand{\thefootnote}{\fnsymbol{footnote}}
\footnotetext[1]{This work was done during an internship in Huawei Noah's Ark Lab.}
\footnotetext[2]{Corresponding author}

Recent progress in vision-language models (VLMs), e.g., CLIP \cite{radford2021learning} and ALIGN \cite{jia2021scaling}, provides a promising opportunity to explicitly leverage human language for addressing downstream recognition tasks efficiently. 
VLMs learn aligned embeddings of image and text via contrastive learning \cite{oord2018representation, chen2020a, he2020momentum}, encouraging the representations of an image and its language description to be similar.   
In the downstream task, providing the task-related content, i.e., the category descriptions, can significantly benefit the recognition via the pre-trained VLM, even to perform zero-shot inference without training samples \cite{radford2021learning}.

\begin{figure}[t]
  \centering
   \includegraphics[width=0.9\linewidth]{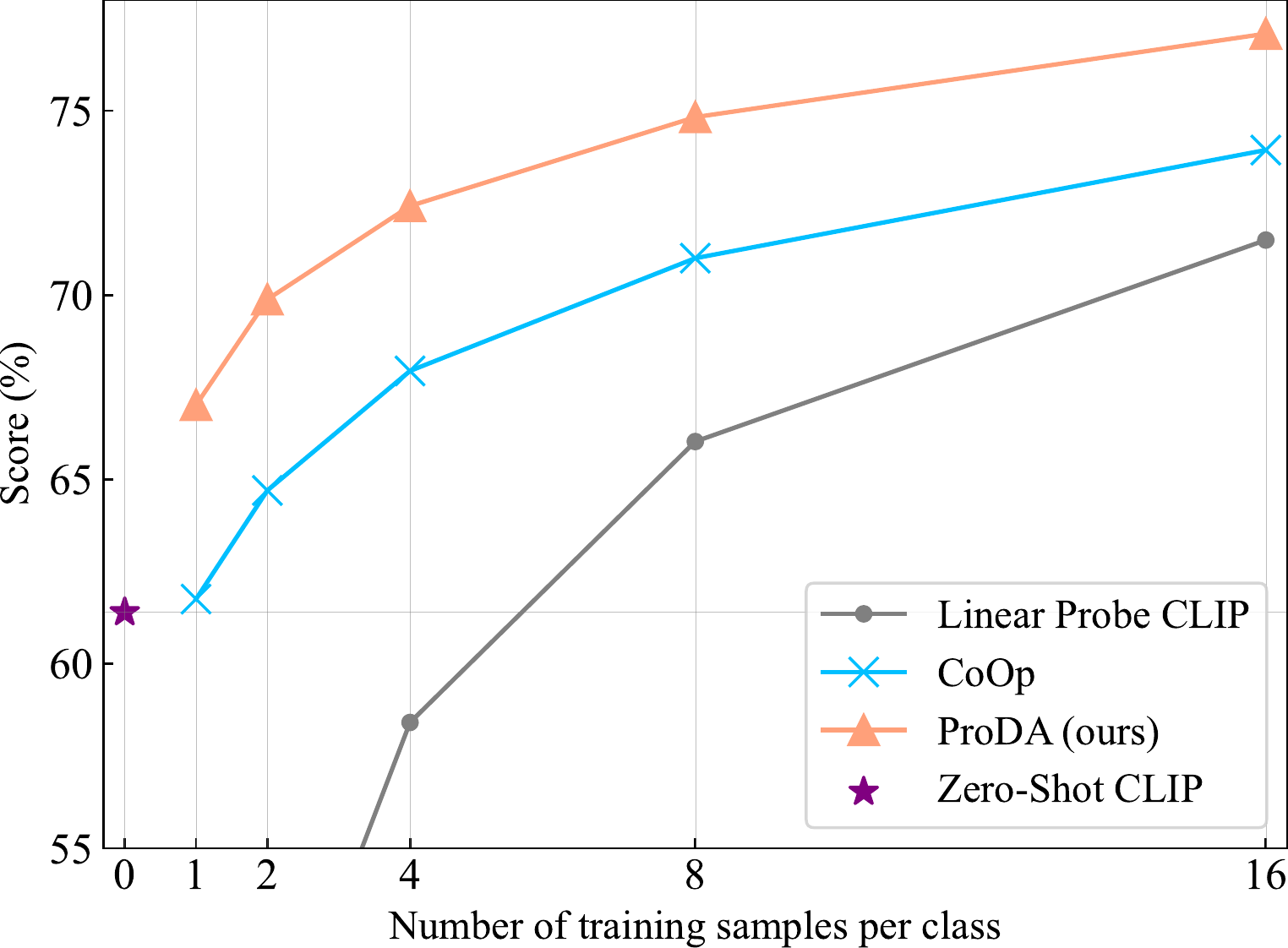}
   \caption{
   Comparison with existing prompt-based methods of leveraging VLM, i.e., hand-crafted prompts (zero-shot CLIP \cite{radford2021learning}) and prompt tuning (CoOp \cite{zhou2021learning}), and the linear probing. We report the average results on $12$ downstream datasets with various training samples. Our method ProDA consistently and substantially outperforms the previous prompt learning approaches. 
   }
   \label{fig:1}
\end{figure}

Leveraging the language, VLMs convert the prior knowledge from humans into exploitable representations to address downstream tasks.
The recognition performance of such methods is highly sensitive to the form of the provided content.
However, it is still a challenging problem to determine the optimal text descriptions.

VLMs \cite{radford2021learning, jia2021scaling} construct category descriptions with the hand-crafted prompt templates.
A default prompt is  ``\texttt{a photo of a \{class\}.}'', which works well for generic object recognition (e.g., on ImageNet \cite{deng2009imagenet} and STL-10 \cite{coates2011an}).
However, it is difficult to handle fine-grained object recognition. On the flower dataset (Oxford Flowers 102 \cite{nilsback2008automated}), a better choice of prompt is ``\texttt{a photo of a \{class\}, a type of flower.}'' \cite{radford2021learning}. In this case, the prompt word ``\texttt{flower}'' indicates the context of the current task, thus providing the more precise description.

From this perspective, the provided text should be adapted to the task-defined context, i.e., \textit{low bias} to the visual representations of the target task. 
However, manually designing inevitably introduces artificial bias and could be sub-optimal for the target task.
Thus, customizing suitable prompts for different recognition tasks relies on repetitive and time-consuming attempts by experts and also requires a large validation set for prompt selection \cite{radford2021learning}.

\begin{figure*}
\centering
\begin{subfigure}{0.4\linewidth}
  \centering
  \includegraphics[width=0.9\linewidth]{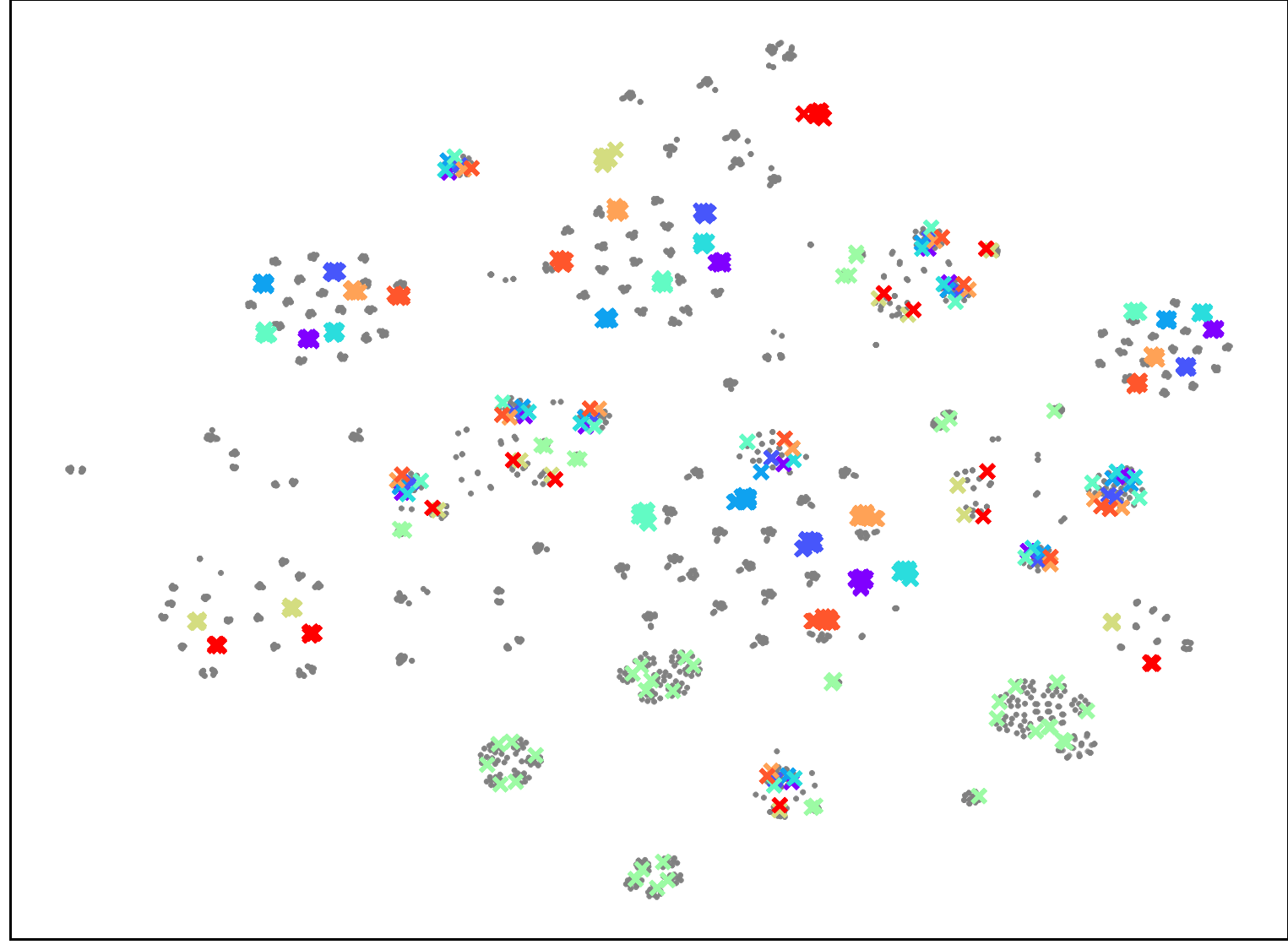}
  \caption{Input embeddings\label{fig:2a}}
  \label{fig:21}
\end{subfigure}%
\begin{subfigure}{.4\linewidth}
  \centering
  \includegraphics[width=0.9\linewidth]{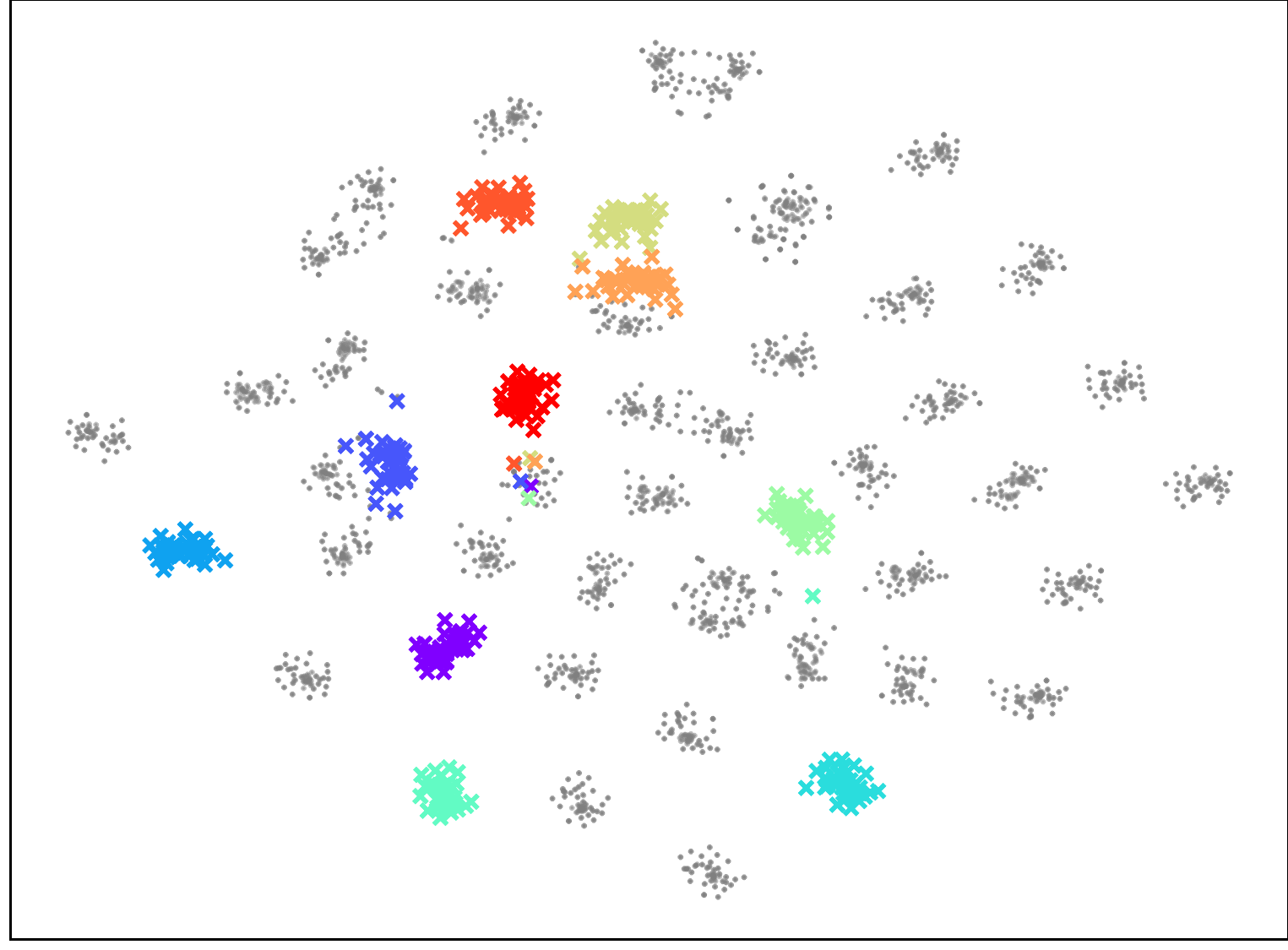}
  \caption{Output embeddings\label{fig:2b}}
  \label{fig:22}
\end{subfigure}
\caption{\textbf{The t-SNE \cite{maaten2008visualizing} visualization of the descriptions for $50$ random categories on ImageNet.}
The descriptions of each category are generated by 80 \textit{hand-crafted} prompts presented by CLIP \cite{radford2021learning}. For clarity, we randomly select $10$ categories and highlight them with different colors. Other categories are in gray. (\textbf{a}) The input embeddings of the text encoder, which are obtained by feeding the raw text into the embedding layer.
Various descriptions within a category are scattered in the space, resulting in difficulty representing their distribution.  
(\textbf{b}) The output embeddings of the text encoder about category descriptions. Relying on the capability of the text encoder, the output embeddings of the descriptions within a category are close to each other, allowing them to be modeled with a simple distribution. (Best viewed in color.)}
\label{fig:2}
\vspace{-0.5cm}
\end{figure*}

Another challenge arises from the diversity of visual content.
Due to inherent factors such as pose, deformation, and lighting condition, there exists significant diversity among various examples within a category \cite{xiao2015the}.
This \textit{intra-class variance} prevents a prompt from sufficiently describing visual variation. 
The prompts are desirable to be \textit{diverse} and \textit{informative}, allowing to handle the variance of visual representations.
Existing work \cite{radford2021learning} ensembles $80$ hand-crafted prompts to predict the categories on ImageNet \cite{deng2009imagenet}, including ``\texttt{a photo of a small \{class\}.}'', ``\texttt{a photo of a large \{class\}.}'', \textit{etc}.
However, it still has the limitation of manual design, requiring cumbersome efforts to select an appropriate but potentially sub-optimal collection of prompts.

We present PROmpt Distribution leArning (ProDA) as a way for automatically learning diverse prompts from data, which can effectively adapt the pre-trained VLM to downstream recognition tasks.
As a data-driven approach, ProDA learns the soft prompts\footnote{Soft prompts, also known as continuous prompts, represent the (word) embeddings of the raw (discrete) prompts.} from a few downstream samples, discovering the task-related content with less bias than manual design. 
Moreover, rather than learning one soft prompt \cite{zhou2021learning}, our ProDA estimates a distribution over diverse and informative prompts to capture the variance of visual representations.
In this way, our approach obtains better generalization to various and unknown samples (Fig. \ref{fig:1}).
Besides, we explicitly differentiate prompts on both construction and semantics to further improve their diversity.

Given the purpose of learning the prompt distribution, the challenge is how to preform the learning efficiently.
Considering the soft prompt is a sequence of tokens (each token is represented by a vector), precise modeling relies on a complicated sequence generation model \cite{brown2020language,oord2016wavenet}, requiring a large number of target samples for training.
In addition, the random nature of prompts leads to the weights of a classification model for the target task being random variables, resulting in the exact computation of the classification loss being intractable (discussed in Sec. \ref{subsec:3.2}).

To address the problem, we adopt an efficient solution which learns the distribution of the output embeddings of the prompts (with class names), i.e., the weights of the target classifier, instead of learning the distribution of the input embeddings of the prompts. 
The underlying intuition is that, although the various descriptions within a category are significantly different in the raw text (or low-level embeddings) (Fig. \ref{fig:2a}), the high-level embeddings of them are usually adjacent (Fig. \ref{fig:2b}), which can be modeled using a simple distribution, such as the multivariate Gaussian distribution in our paper. 
Moreover, based on the Gaussian distribution assumption, we propose a surrogate objective, an upper bound of the original optimization objective, for effective training, avoiding the intractable calculation.

We conduct large-scale experiments on 12 datasets to demonstrate the effectiveness of our method, which has a consistent and significant improvement over existing baselines. For example, ProDA with 1 sample per category relatively improves the average result by $9.1\%$ compared to the human-crafted prompts. 

\section{Related Work}
\label{sec:relwork}

\paragraph{Vision-Language Pre-Trained Models.}
A promising way to build a transferable and usable recognition model is vision-language pre-training, which learns the connection between image content and language. A lot of approaches attempt to learn representations by predicting the captions of images \cite{joulin2016learning, li2017learning, sariyildiz2020learning, desai2021virtex, zhang2021contrastive}. The main obstacle of them is the size of training data. The models are train on relatively small datasets (e.g., Flickr \cite{joulin2016learning} and COCO Captions \cite{desai2021virtex}), limiting their performances. Recently, VLMs based on contrastive learning have demonstrated impressive results by leveraging web-scale noisy image-text pairs. These methods, CLIP \cite{radford2021learning} and ALIGN \cite{jia2021scaling}, learn the aligned representations of image and text by the contrastive loss, which pulls the representations of matching image-text pairs close and pushes those of mismatching pairs far away. Based on natural language supervision, these VLMs not only learn powerful visual representations but are also easily transferred to various downstream tasks.

\vspace{-0.3cm}

\paragraph{Prompt Learning.}
Prompt learning/engineering stems from recent advances in natural language processing (NLP). A novel prompt-based paradigm \cite{petroni2019language, brown2020language, shin2020autoprompt, jiang2020how, li2021prefix, schick2021it, lester2021the} for exploiting pre-trained language models has gradually replaced the traditional transfer approach of fine-tuning \cite{radford2018language, dong2019unified} in NLP. 
The main idea of prompt learning is to formalize various NLP tasks to masked language modeling problems, which is similar to the pre-training of language models \cite{devlin2018bert, radford2019language,radford2021learning}, by adopting different prompt templates.
Discovering the appropriate prompt is central to this line of works.
The preliminary works \cite{radford2019language, petroni2019language, brown2020language} elaborately design human-crafted prompts, which is known as \textit{prompt engineering}. 
Since manual design is sensitive and difficult, a series of approaches \cite{jiang2020how,shin2020autoprompt} focus on \textit{automatically} generating desired (discrete) prompts in the natural language space. 
Recently, some works \cite{li2021prefix, zhong2021factual, han2021ptr,lester2021the}, also known as \textit{prompt tuning}, attempt to learn soft (continuous) prompts directly instead of searching for discrete prompts.

While prompt learning receives considerable attention in NLP, it remains underexplored in computer vision. Pre-trained VLMs \cite{radford2021learning, jia2021scaling} introduce hand-crafted prompts to perform zero-shot inference on the downstream tasks. 
A concurrent work (CoOp \cite{zhou2021learning}) adopts the prompt tuning approach of NLP, which learns a soft prompt via minimizing the classification loss on the target task.
CoOp is similar to our approach in the sense that both works learn the prompt(s) in a data-driven manner.
However, learning a single prompt \cite{zhou2021learning} neglects the diversity of visual representations, which is challenging to capture various changes of visual content.
In contrast, our method learns the distribution of diverse prompts, resulting in better generalization. We relatively improve the average results by $8.5\%$ compared to CoOp in the $1$-shot setting on $12$ datasets.

\section{Method}
\label{sec:method}

In this section, we present prompt distribution learning (ProDA), which effectively adapts a pre-trained VLM to various downstream visual recognition tasks.
Without loss of generality, we adopt the public implementation of CLIP \cite{radford2021learning} as our pre-trained model.

\subsection{Preliminaries}
\label{sec:pre}

The VLM consists of an \textit{image encoder} $f(\cdot)$ and a \textit{text encoder} $g(\cdot)$. 
We denote $\mathbf{z}$$=$$f(\mathbf{x})/||f(\mathbf{x})||_2$ and $\mathbf{w}=g(\mathbf{t})/||g(\mathbf{t})||_2$, which are the normalized output embeddings of the image $\mathbf{x}$ and the text $\mathbf{t}$, respectively.
Notice that $\mathbf{t}$ is the input embedding (of the text encoder), which is obtained by feeding the raw text to an embedding layer.
During pre-training, CLIP trains the image and text encoders on massive image-text pairs by the contrastive loss, which considers matching image-text pairs as positive and mismatching pairs as negative. 
\vspace{-0.3cm}

\begin{figure*}[ht]
  \centering
  \includegraphics[width=0.9\linewidth]{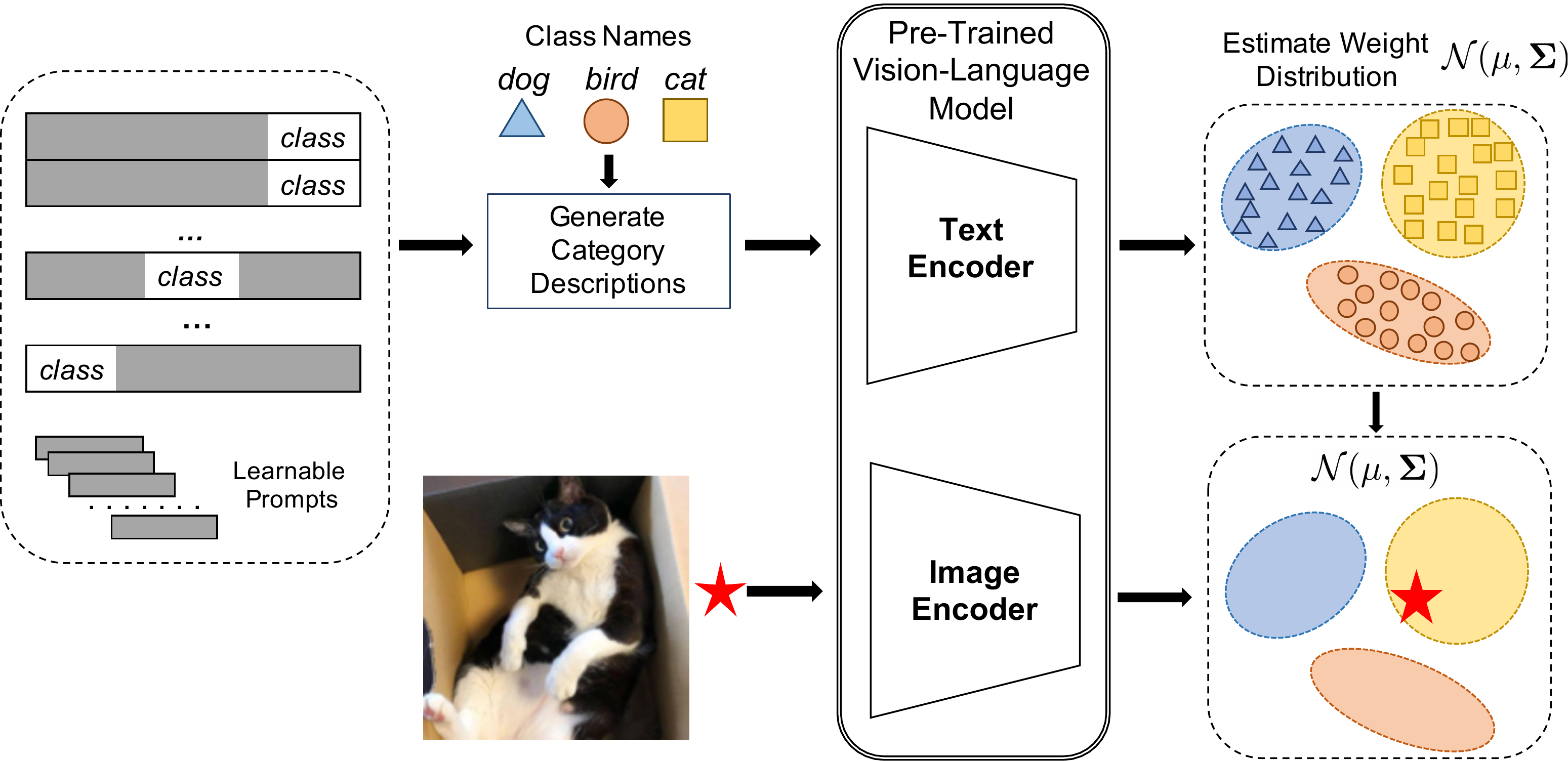}
   \caption{\textbf{Overview of the architecture of ProDA.} The class names and various learnable prompts are integrated to generate diverse category descriptions on the downstream recognition task. The output embeddings of these descriptions as the weights of linear classifiers are used to estimate the weight distribution.
Given the weight distribution, we can minimize the empirical classification error and predict the classes of test samples.}
   \label{fig:3}
\end{figure*}

\paragraph{Prompt Design.} Given the pre-trained models $f(\cdot)$ and $g(\cdot)$, CLIP \cite{radford2021learning} performs the zero-shot inference on a downstream recognition task by manually designing the prompt template. Given the class names of the downstream task, the category descriptions $\{\mathbf{t}_c\}^C_{c=1}$ are generated with the predefined prompt, such as ``\texttt{a photo of a \{class\}.}'', where $C$ is the number of classes.
Then, we can predict the class of the test sample $\textbf{x}$ with the prediction probability as:

\begin{footnotesize}
\begin{equation}
   p(y|\mathbf{x}) = \frac{ e^{\mathbf{z}^{T} \mathbf{w}_{y}/\tau} }{ \sum_{c=1}^{C} e^{\mathbf{z}^{T} \mathbf{w}_{c}/\tau} },
   \label{eq:1}
\end{equation}
\end{footnotesize}

\noindent where $\textbf{z}\in \mathbb{R}^d$ and $\textbf{w}_c\in\mathbb{R}^d$ are the normalized embeddings of the image $\textbf{x}$ and text $\textbf{t}_c$, respectively, $d$ is the dimensionality of the output embedding, and $\tau$ is the temperature\footnote{CLIP learns $\tau$ in the pre-training. We fix $\tau$ in the downstream tasks.}. $\mathbf{w}_{1:C}$$=$$[\textbf{w}^T_1,...,\textbf{w}^T_C]^T\in\mathbb{R}^{dC}$ can be considered as the weights of a linear classifier, which are used to classify the image $\mathbf{x}$.

\vspace{-0.3cm}

\paragraph{Prompt Tuning.} An alternative way to generate the weight $\mathbf{w}_{1:C}$ of the target classifier is \textit{prompt tuning}, which learns a suitable prompt from a few samples on the target task.
Prompt tuning is originally proposed to probe pre-trained language models \cite{lester2021the}. 
Recently, a concurrent work \cite{zhou2021learning} uses it to learn an appropriate prompt for VLMs instead of manually designing it.

Given a learnable continuous prompt $\mathbf{P}\in\mathbb{R}^{p \times e}$ with random initialization, where $p$ is the number of tokens and $e$ is the dimensionality of input (word) embedding, the description of each class $\mathbf{t}_c(\mathbf{P})$ is obtained via concatenating the embeddings of each class name and the prompt $\mathbf{P}$. 
Then, with the generated weight vector $\mathbf{w}_{1:C}(\mathbf{P})=[\mathbf{w}^T_1(\mathbf{P}),...,\mathbf{w}^T_C(\mathbf{P})]^T$, where $\mathbf{w}_c(\mathbf{P})=g(\mathbf{t}_c(\mathbf{P}))$, we can learn the prompt $\mathbf{P}$ with a few training samples $\mathcal{D}^{tr}=\{(\mathbf{x}_i, y_i)\}_{i=1}^{M}$ by minimizing the following objective:

\begin{footnotesize}
\begin{align}
  \mathcal{L}(\mathbf{P}) &= \mathop{\mathbb{E}}\limits_{\mathbf{x}_i,y_i} \left[  -\log p(y_i|\mathbf{x}_i, \mathbf{w}_{1:C}(\mathbf{P})) \right] \\
                &= \mathop{\mathbb{E}}\limits_{\mathbf{x}_i,y_i} \left[ -\log \frac{ e^{\mathbf{z}_i^{T} \mathbf{w}_{y_i}(\mathbf{P})/\tau} }{ \sum_{c=1}^{C} e^{\mathbf{z}_i^{T} \mathbf{w}_{c}(\mathbf{P})/\tau} } \right],
\end{align}
\end{footnotesize}

\noindent where $\mathbf{z}_i$ is the normalized embedding of $\mathbf{x}_i$. Notice that all parameters of the pre-trained model are frozen during the learning process.
After learning, prompt tuning leverages the learned prompt $\mathbf{P}$ to generate the target classifier and classify test samples.

\subsection{Learning the Prompt Distribution}
\label{subsec:3.2}

In order to handle diverse visual variations, our approach ProDA aims to learn the distribution of various prompts.
Intuitively, we should learn an optimal prompt distribution $p(\mathbf{P})$, which minimizes the empirical classification loss.
In this case, the classifier weights $\mathbf{w}_{1:C}(\mathbf{P})$ follow a distribution determined by $p(\mathbf{P})$ and the text encoder $g(\cdot)$, resulting in the prediction probability $p(y|\mathbf{x})$ to be a \textit{marginal likelihood} $\mathbb{E}_\mathbf{P}[p(y|\mathbf{x},\mathbf{w}_{1:C}(\mathbf{P}))]$.
Unfortunately, explicitly computing this marginal likelihood is intractable, which requires the integration over $\mathbf{P}$.
In a special case where $\mathbf{P}$ is a discrete random variable, the computing is possible. 
However, it restricts the learning for the overall prompts.
Moreover, learning the exact distribution of prompts is difficult, requiring a complicated sequence generation model \cite{brown2020language,oord2016wavenet}.

In this work, we propose an efficient method to indirectly learn the prompt distribution by learning the distribution of the classifier weights, i.e., the output embeddings of the category descriptions. 
Although the original distribution of prompt $\mathbf{P}$ is complex, the generated weights $\mathbf{w}_c(\mathbf{P})$ within a category are adjacent, as shown in Fig. \ref{fig:22}, which can be modeled with the multivariate Gaussian distribution.
Recent works \cite{wang2021regularizing,9582738,liu2022category,zhang2022adversarial} show that the Gaussian distribution is effective to model the representations learned by neural networks. 

Specifically, we assume $\mathcal{N}(\mathbf{\mu}_{1:C}, \mathbf{\Sigma}_{1:C})$ is the ``true'' distribution of the weights $\mathbf{w}_{1:C}$.
We maintain a collection of learnable continuous prompts $\mathcal{P}^K \triangleq \{\mathbf{P}_k\}^K_{k=1}$.
The mean and covariance of the ``true'' weight distribution can be estimated from a series of classifier weights $\{\mathbf{w}_{1:C}(\mathbf{P}_k)\}_{k=1}^K$, which are generated by the prompts from $\mathcal{P}^K$.
Fig. \ref{fig:3} illustrates the architecture of our model. Next, we propose a surrogate loss for efficient training.

\paragraph{Optimization.} 
Learning the weight distribution relies on learning an optimal prompt collection $\mathcal{P}^K$.
Given the weights of the $K$ classifiers $\{\mathbf{w}_{1:C}(\mathbf{P}_k)\}^K_{k=1}$, we can estimate the mean $\mathbf{\mu}_{1:C}(\mathcal{P}^K)$ and covariance matrix $\mathbf{\Sigma}_{1:C}(\mathcal{P}^K)$.
The prompt collection is trained by minimizing the empirical classification loss:

\begin{footnotesize}
\begin{align}
    \mathcal{L}(\mathcal{P}^K) &= \mathop{\mathbb{E}}\limits_{\mathbf{x}_i,y_i} \left[ - \log \mathop{\mathbb{E}}\limits_{\mathbf{w}_{1:C}} p(y_i|\mathbf{x}_i, \mathbf{w}_{1:C}) \right]\\
    &= \mathop{\mathbb{E}}\limits_{\mathbf{x}_i,y_i} \left[ - \log \mathop{\mathbb{E}}\limits_{\mathbf{w}_{1:C}} \frac{e^{\mathbf{z}_i^T \mathbf{w}_{y_i} / \tau}}{\sum_c e^{\mathbf{z}_i^T \mathbf{w}_c /\tau }} \right]
\end{align}
\end{footnotesize}

\noindent where $\mathbf{w}_{1:C}\sim \mathcal{N}(\mathbf{\mu}_{1:C}(\mathcal{P}^K), \mathbf{\Sigma}_{1:C}(\mathcal{P}^K))$.

However, even with the Gaussian distribution assumption, the exact computation of the marginal likelihood is still intractable in the multi-class case \cite{rasmussen2005gaussian,williams1998bayesian}. To address this problem, we derive an upper bound of the loss for efficient optimization.

\begin{proposition}
Suppose that $\mathbf{w}_{1:C}=[\mathbf{w}^T_1,..,\mathbf{w}^T_C]^T\in\mathbb{R}^{dC}$ follows $\mathcal{N}(\mathbf{\mu}_{1:C}(\mathcal{P}^K), \mathbf{\Sigma}_{1:C}(\mathcal{P}^K))$. Let $\mathbf{\Sigma}_{ij}(\mathcal{P}^K)$ be the covariance matrix of $\mathbf{w}_i$ and $\mathbf{w}_j$, $\mathbf{\mu}_{i}(\mathcal{P}^K)$ be the mean of $\mathbf{w}_i$, and $\mathbf{A}_{i,j}=\mathbf{\Sigma}_{ii} + \mathbf{\Sigma}_{jj}-\mathbf{\Sigma}_{ij}-\mathbf{\Sigma}_{ji}$. Then it holds that

\begin{footnotesize}
\begin{align}
    \mathcal{L}(\mathcal{P}^K) &\leq \mathop{\mathbb{E}}\limits_{\mathbf{x}_i,y_i} \left[ - \log \frac{e^{\mathbf{z}_i^T \mathbf{\mu}_{y_i}(\mathcal{P}^K)/\tau}}{\sum_c e^{ \mathbf{z}_i^T \mathbf{\mu}_c(\mathcal{P}^K)/\tau + \mathbf{z}^T_i \mathbf{A}_{c,y_i} \mathbf{z}_i / 2\tau^2} } \right]\\
    &\triangleq \mathcal{L}_{upper}(\mathcal{P}^K).
\label{equ:7}
\end{align}
\end{footnotesize}
\end{proposition}
The proof is provided in our supplementary materials. By minimizing $\mathcal{L}_{upper}$, our method efficiently trains the prompt collections for estimating the weight distribution, which is used to predict the classes of test image samples.

\paragraph{Inference.}
Given the learned prompts $\mathcal{P}^K$, the classifier weights $\mathbf{w}_{1:C}$ follow $\mathcal{N}(\mathbf{\mu}_{1:C}(\mathcal{P}^K), \mathbf{\Sigma}_{1:C}(\mathcal{P}^K))$. The class of a test sample is predicted by the prediction probability $\mathbb{E}_{\mathbf{W}_{1:C}} [p(y|\mathbf{x},\mathbf{w}_{1:C})]$. Although the explicit calculation is intractable, some numerical approximations can be used in the inference. A straightforward approach is Monte Carlo \cite{williams1998bayesian,rasmussen2005gaussian}, which requires sampling multiple classifier weights, but  it results in increased inference computation. In our experiments, we find that simply using the mean of the weight distribution for classification works well, i.e., predicting by $p(y|\mathbf{x},\mathbb{E}(\mathbf{w}_{1:C}))$. It also allows our method to have no additional computational overhead for inference.

\subsection{Improving Prompt Diversity}
Since the parameters of the weight distribution are estimated from the prompt collection, the quality of the prompts affects the obtained distribution.
Diverse prompts can describe the visual content more sufficiently, improving the generalization on the test samples. 
The work \cite{zhang2020the} demonstrates that diverse classifiers are able to enhance generalization. 
To further improve the diversity of prompts, we explicitly differentiate the prompts of $\mathcal{P}^K$.

\paragraph{Position Diversity.}
A common way to combine the prompt and the category name is to put the category name at the end of the prompt. However, the generated text descriptions are biased. To improve the diversity of the generated text descriptions, we insert the category name in the front, middle, and end positions of different prompts. In our experiments, the proportions of these three types on 
$\mathcal{P}^K$ are 1/4, 1/4, and 1/2.

\paragraph{Semantic Orthogonality.}
Different prompts should represent different contents. A natural way is to encourage them to have dissimilar semantics. We feed the prompts $\{\mathbf{P}_k\}_{k=1}^K$ without incorporating the category name into the pre-trained text encoder to obtain their semantic embeddings $\{g(\mathbf{P}_k)\}_{k=1}^K$. The following semantic orthogonality loss is use to encourage the prompts to be dissimilar:

\begin{footnotesize}
\begin{equation}
    \mathcal{L}_{so}(\mathcal{P}^K) = \frac{1}{K(K-1)}\sum_{i=1}^K \sum_{j=i+1}^K |<g(\mathbf{P}_i), g(\mathbf{P}_j)>|,
\label{eq:8}
\end{equation}
\end{footnotesize}

\noindent where $<\cdot,\cdot>$ denotes the cosine similarity. Then the total training loss is:
\begin{equation}
    \mathcal{L}=\mathcal{L}_{upper} + \lambda \mathcal{L}_{so},
\label{eq:loss_all}
\end{equation}

\noindent where $\lambda$ is a hyper-parameter. We set $\lambda=0.1$ for all experiments.

\begin{algorithm}[H]
\caption{Pseudocode of ProDA Training.}
\label{alg:code}
\begin{algorithmic}[1]
    \STATE {\bfseries Require:} The pre-trianed VLM encoders of image $f$ and text $g$
    \STATE {\bfseries Require:} The training set $\mathcal{D}^{tr}$ of the target task
    \STATE {\bfseries Require:} The input word embeddings of class names $\{\mathbf{e}_c\}^C_{c=1}$
    \STATE {Randomly initialize the prompt collection $\mathcal{P}^K$}
    \FOR{$t=0$ {\bfseries to} $T$}
    \STATE Sample a mini-batch $\{( \mathbf{x}_{i}, {y_{i}} )\}^{B_x}_{i=1}$  from $\mathcal{D}^{tr}$
    \STATE Compute $\mathbf{z}_i=f(\mathbf{x}_i)$, $i$$=$$1,..,B_x$
    \STATE Sample a mini-batch $\{ \mathbf{P}_{b}\}_{b=1}^{B_P}$ from $\mathcal{P}^K$
    \STATE Combine the $\mathbf{P}_{b}$ and class name $\mathbf{e}_c$ to generate class description $t_c(\mathbf{P}_b)$, $c$$=$$1,..,C$;    $b$$=$$1,..,B_P$
    \STATE Compute $\mathbf{w}_c(\mathbf{P}_b)$$=$$g(t_c(\mathbf{P}_b))$, $c$$=$$1,..,C$; $b$$=$$1,$ $..,B_P$
    \STATE Let $\mathbf{w}_{1:C}(\mathbf{P}_b)=[\mathbf{w}^T_1(\mathbf{P}_b),...,\mathbf{w}^T_C(\mathbf{P}_b)]^T$
    \STATE Compute the mean $\mathbf{\mu}$ and covariance martix $\mathbf{\Sigma}$ of $B_P$ vectors $\{\mathbf{w}_{1:C}(\mathbf{P}_b)\}^{B_P}_{b=1}$ 
    \STATE Compute $\mathcal{L}_{upper}$
    according to Eq. (\ref{equ:7})
    \STATE Compute $\mathcal{L}_{so}$ according to Eq. (\ref{eq:8})
    \STATE Compute the total loss $\mathcal{L}$ according to Eq. (\ref{eq:loss_all})
    \STATE Update $\{ \mathbf{P}_{b}\}_{b=1}^{B_P}$ by gradient descent
    \ENDFOR
\end{algorithmic}
\end{algorithm}

\begin{figure*}[tbp]
  \centering
  \includegraphics[width=1.0\linewidth]{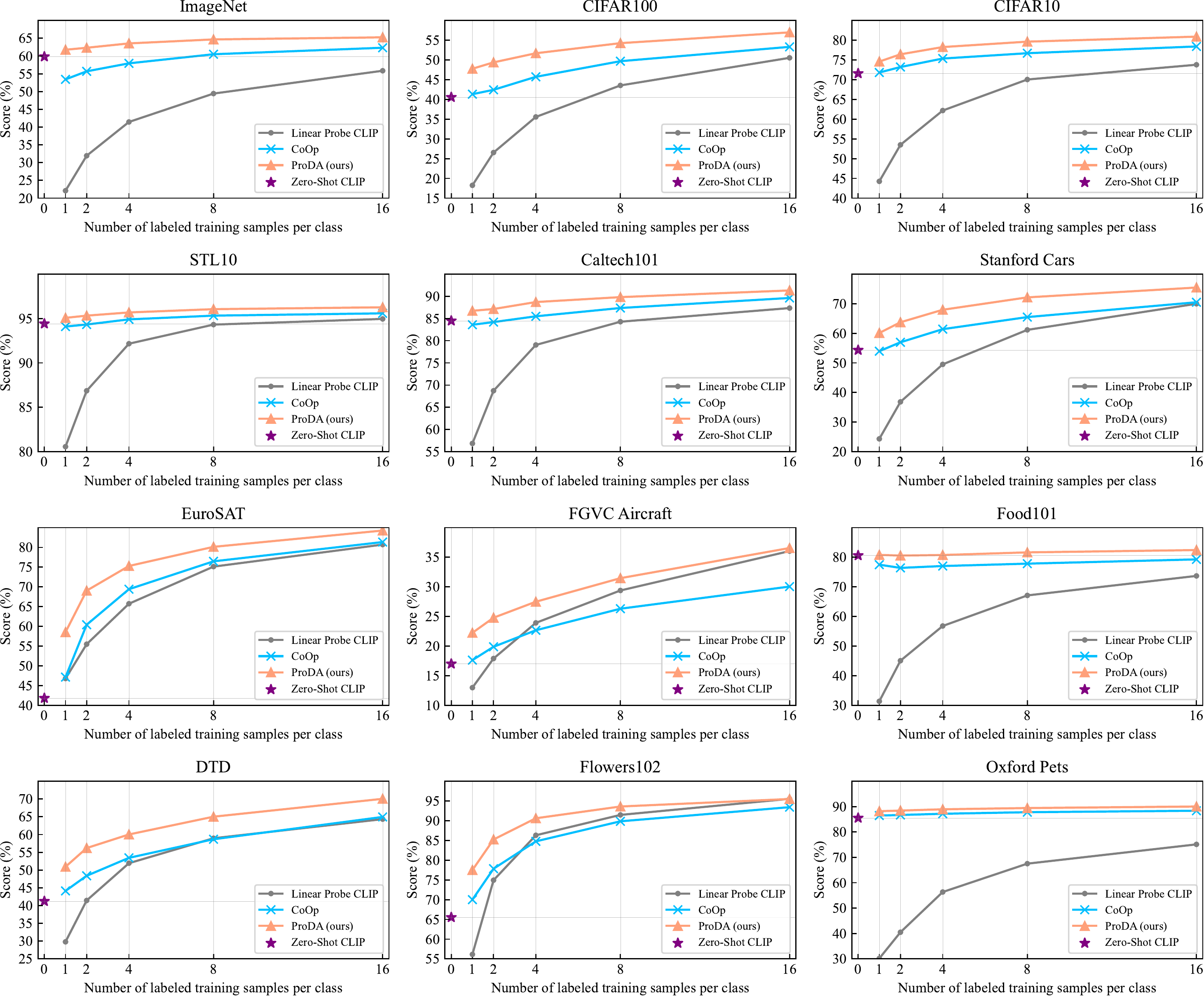}
   \caption{Comparison with two prompt-based methods and the linear probe on various downstream tasks. Our method consistently and significantly outperforms these methods.}
   \label{fig:detail_res}
\end{figure*}

\subsection{Implementation}

Unless otherwise specified, we adopt the publicly available CLIP model with the ResNet-50 \cite{he2016deep} visual backbone as our pre-trained model ($d$$=$$1024$).
To reduce memory consumption, we randomly sample a batch of prompts $\{\mathbf{P}_b\}^B_{b=1}$ from the prompt collection in each training iteration, instead of using all the prompts.
These $B$ prompts and $C$ category names are coupled to generate $B\times C$ category descriptions, which are used to form $B$ classifiers for estimating the distribution of classifier weights.
Then we can minimize Eq. \ref{eq:loss_all} on these $B$ prompts.
In inference, all prompts of the collection are used to estimate the distribution of the classifier weights.
Besides, we approximate the covariance matrix $\mathbf{\Sigma}_{i,j}$ with the diagonal matrix to further reduce memory consumption.
In this way, we train our model with $1$ GPU on most datasets.
On ImageNet, we adopt $4$ GPU to accelerate the training. 
Algorithm \ref{alg:code} provides the pseudo-code of the training procedure.


\section{Experiments}
\label{sec:exper}

\paragraph{Datasets.} 
We evaluate our ProDA on $12$ downstream classification datasets, including general object recognition (ImageNet-1k \cite{deng2009imagenet}, CIFAR-10 \cite{krizhevsky2009learning}, CIFAR-100 \cite{krizhevsky2009learning}, STL-10 \cite{coates2011an}, and Caltech-101 \cite{fei-fei2007learning}), fine-grained object recognition (Oxford-IIIT Pets \cite{parkhi2012cats}, Food-101 \cite{bossard2014food}, Stanford Cars \cite{krause20133d}, Oxford Flowers 102 \cite{nilsback2008automated}, and FGVC Aircraft \cite{maji13fine-grained}), remote sensing recognition (EuroSAT \cite{helber2019eurosat}), and texture recognition (DTD \cite{cimpoi2014describing}). The details and evaluation metrics of each dataset are provided in the supplementary materials. 

\paragraph{Training Details}
The number of tokens in each prompt and the number of prompts in the collection are set to 16 and 32, respectively. 
The batch size of prompts is $4$. 
We train the prompts for $100$ epochs with SGD optimizer.
The momentum of SGD is $0.9$.
We set the learning rate using the linear scaling rule $lr\times$ImageBatchSize$/5$, with a base $lr = 0.001$.
The batch size of images is $20$ on most datasets.
We use the larger batch size $100$ on ImageNet.
The learning rate has a cosine decay schedule.
We use the model of the last training epoch for evaluation. 
\vspace{-0.3cm}

\begin{figure*}[htbp]
    \centering
  \subfloat[ProDA vs Hand-Crafted Prompt\label{fig:bar_1}]{%
       \includegraphics[width=0.5\linewidth]{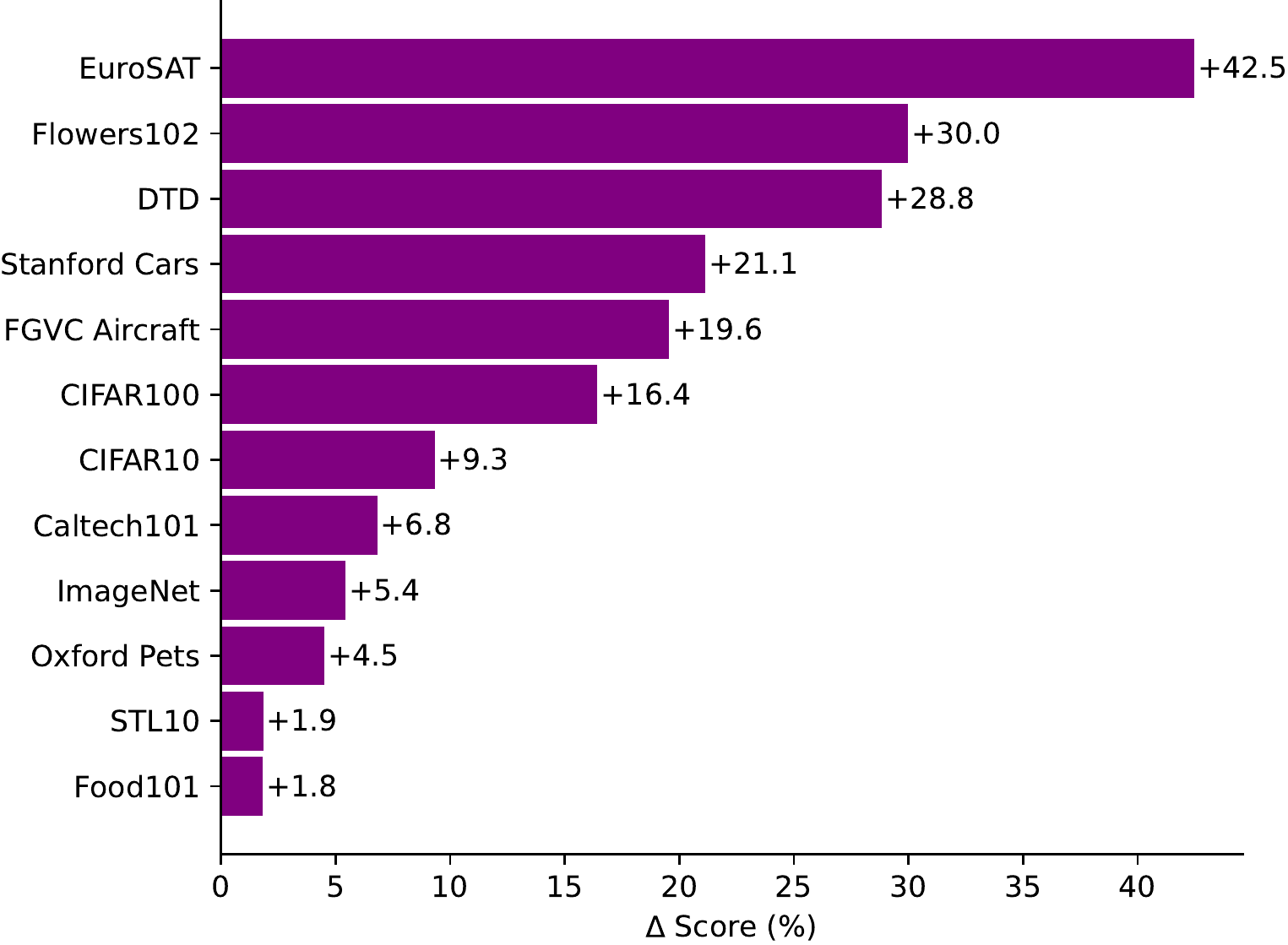}}
    \hfill
  \subfloat[ProDA vs CoOp\label{fig:bar_1}]{%
        \includegraphics[width=0.5\linewidth]{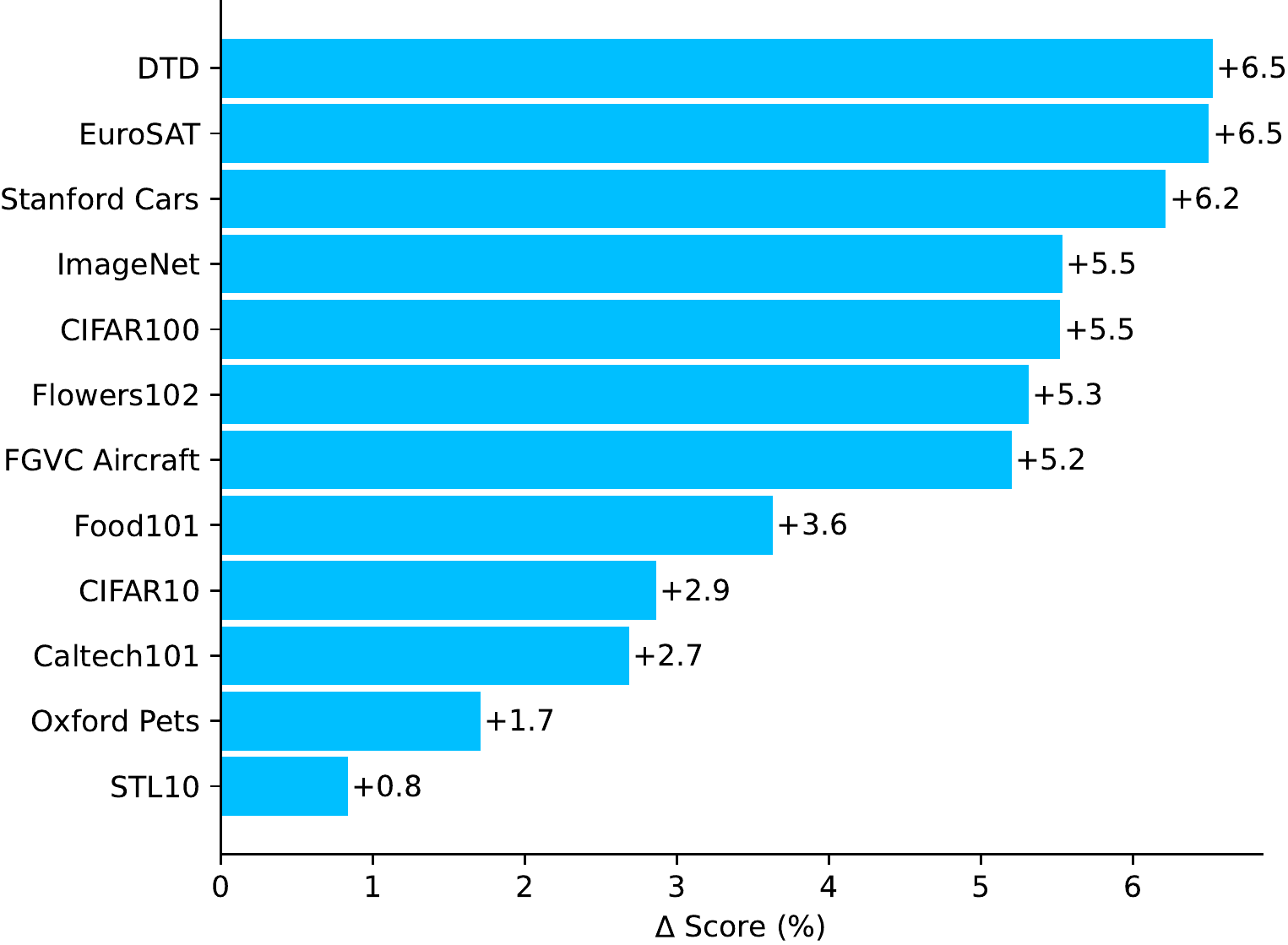}}
    \hfill
  \caption{\textbf{Comparison with prompt-based methods.} We show the \textit{absolute improvement} of our approach compared to hand-crafted prompts \cite{radford2021learning} and prompt tuning (CoOp \cite{zhou2021learning}) on each downstream task. (\textbf{a}) We compare the hand-crafted prompts to our ProDA with $16$ samples per category. (\textbf{b}) Our method is compared with the prompt tuning by their average results 
 of various shots (1, 2, 4, 8, and 16).}
  \label{fig:bar} 
\end{figure*}

\paragraph{Baselines.} 
We compare our approach with two existing prompt-based methods (zero-shot CLIP and prompt tuning) and the linear probe CLIP. The \textbf{zero-shot CLIP} \cite{radford2021learning} uses hand-crafted prompts to generate the target classifier on the downstream task, as discussed in Sec. \ref{sec:pre}. The prompt templates applied in each dataset are the same as CLIP \cite{radford2021learning}. 
We note that CLIP uses the full validation set of each dataset, which often has thousands of samples, to manually design these prompts \cite{radford2021learning}.
In addition, following the guideline of CLIP, we ensemble multiple classifiers for improving the performance of zero-shot CLIP. The prompt tuning  (\textbf{CoOp} \cite{zhou2021learning}) learns a soft prompt by minimizing the classification loss, which is discussed in Sec. \ref{sec:pre}. Our implementation obtains slightly better results than those reported in CoOp \cite{zhou2021learning}. 
For \textbf{linear probe CLIP}, we train a logistic regression classifier on the features of training images. Existing work demonstrates that training a linear classifier on the embeddings of the pre-trained model is a strong baseline for few-shot learning \cite{tian2020rethinking}.
Details of the baseline methods are provided in the supplementary materials.

\paragraph{Evaluation protocols.} 
We follow the few-shot transfer setting on CLIP \cite{radford2021learning}, which learns with 1, 2, 4, 8, and 16 labeled samples per class on each downstream task. Training examples are sampled from the training set of each dataset. After training, each method is evaluated on the full test set of the downstream task with the corresponding metric. We report the average results over $3$ runs.

\subsection{Main Results}

Fig. \ref{fig:detail_res} shows the comparison with the baseline methods on $12$ downstream tasks. More detailed results are provided in the supplementary materials.
The average results over all datasets are given in Fig. \ref{fig:1}. We also provide a summary of the absolute improvement of our approach compared to the two prompt-based methods in Fig. \ref{fig:bar}.
All methods adopt the same pre-trained CLIP model. 

In comparison to the hand-crafted prompts (zero-shot CLIP), our approach substantially improves the performance. Our ProDA relatively improves the average results by $9.1$\% with $1$ training sample per class and $25$\% with $16$ training samples per class. In the uncommon datasets such as EuroSAT and DTD, the relative improvements are more significant ($40$\% and $25$\% in the $1$-shot setting). 
We consider that, for these special images (remote sensing or texture images), selecting prompts based on human experience would be more difficult and introduce more artificial bias. 
These results support our motivation of learning the low-bias prompt distribution.

In addition, our approach consistently and significantly outperforms CoOp. We have $8.5$\% relative average performance improvement in $1$-shot and $4.3$\% in $16$-shot compared with it. 
These results suggest the necessary of learning the distribution of diverse prompts for handling the variance of visual contents.

The comparison with linear probe CLIP demonstrates the benefit of using category names for recognition in the few-shot setting. In $1$-shot, our approach has relatively $77$\% higher average score than the linear probe CLIP ($67.0$\% vs $37.8$\%). 
Natural language provides dense task-related information rather than images. 
Our results indicate that prompt learning is an efficient way to address vision tasks.

Overall, our method ProDA substantially outperforms its prompt learning/engineer counterparts. These results demonstrate the effectiveness of our approach, which learns a low-bias and diverse prompt distribution. They also indicate that leveraging natural language to provide the task-related content can be a promising paradigm to address downstream recognition tasks efficiently.

\subsection{Ablation Study}

In this section, we ablate the different components in our proposed ProDA.
\vspace{-0.2cm}

\begin{table}[tbp]
\footnotesize
\centering
\begin{tabular}{c|ccccc}
 & \multicolumn{5}{c}{$\#$ of training samples per class} \\
 & 1 & 2  & 4 & 8  & 16 \\
\hline
CoOp \cite{zhou2021learning} & 61.8 & 64.7 & 67.9 & 71.0 & 73.9 \\
\hline
Ours w/o $\mathcal{L}_{upper}$ &  65.8 & 68.8 & 71.6 & 74.2 & 76.6 \\
Ours w/o pos. div. & 66.6 & 69.4  &  71.8  &  74.3  &  76.6 \\
Ours w/o sem. orth. & 66.8 &  69.6 &   72.2 &   74.5 & 76.8 \\
Ours & \textbf{67.0} & \textbf{69.9} & \textbf{72.4} & \textbf{74.8} & \textbf{77.1} \\
\Xhline{1.0pt}
\end{tabular}

\caption{\textbf{Ablation study} of our ProDA approach. We show the average scores on $12$ downstream tasks of various training samples. w/o $\mathcal{L}_{upper}$: compute the standard classification loss by treating the mean of the classifier weights as an ensemble weight, without learning the weight distribution; w/o pos. div.: all prompts are combined with the category name at the end; w/o sem. orth.: the semantic orthogonal loss $\mathcal{L}_{so}$ is not used.}
\label{exp:ablation}
\end{table}

\begin{table}[tbp]
\centering
\footnotesize
\begin{tabular}{l|c|ccccc}
& & \multicolumn{5}{c}{$\#$ of training samples per class} \\
 & text bsz & 1 & 2  & 4 & 8  & 16 \\
\hline
mini-batch prompts& 1$\times$ &  \textbf{74.6} & \textbf{76.4} &	\textbf{78.3}	& 79.6 &	80.9\\
all prompts& 8$\times$ & 74.4 &	76.3 & \textbf{78.3} & \textbf{79.7} & \textbf{81.0}\\
\Xhline{1.0pt}
\end{tabular}
\caption{\textbf{Sampling mini-batch prompts.} We compare the sampling strategy with using all prompts on CIFAR-10 \cite{krizhevsky2009learning}. The ``text bsz'' denotes the batch size of input texts to the text encoder in each training forward.}
\label{exp:minibatch}
\vspace{-0.3cm}
\end{table}

\paragraph{Weight distribution.} 
Another way to learn diversity prompts is to aggregate the classifiers generated by multiple prompts and optimize the prompts using standard classification losses. Table \ref{exp:ablation} shows the comparison of our approach and this strategy. Our method consistently outperforms it, demonstrating the effectiveness of learning the weight distribution. We find that this strategy is also significantly better than prompt tuning \cite{zhou2021learning}, which supports our motivation of learning diverse prompts to capture the visual content.
\vspace{-0.3cm}

\paragraph{Diversity constraint.} Table \ref{exp:ablation} shows the effect of the prompt diversity constraints on recognition performance. Encouraging the prompts with diverse positions improve the average scores. In addition, constraining prompt semantics orthogonally also slightly improves the performance. 
\vspace{-0.3cm}

\begin{figure}[tbp]
  \centering
   \includegraphics[width=0.9\linewidth]{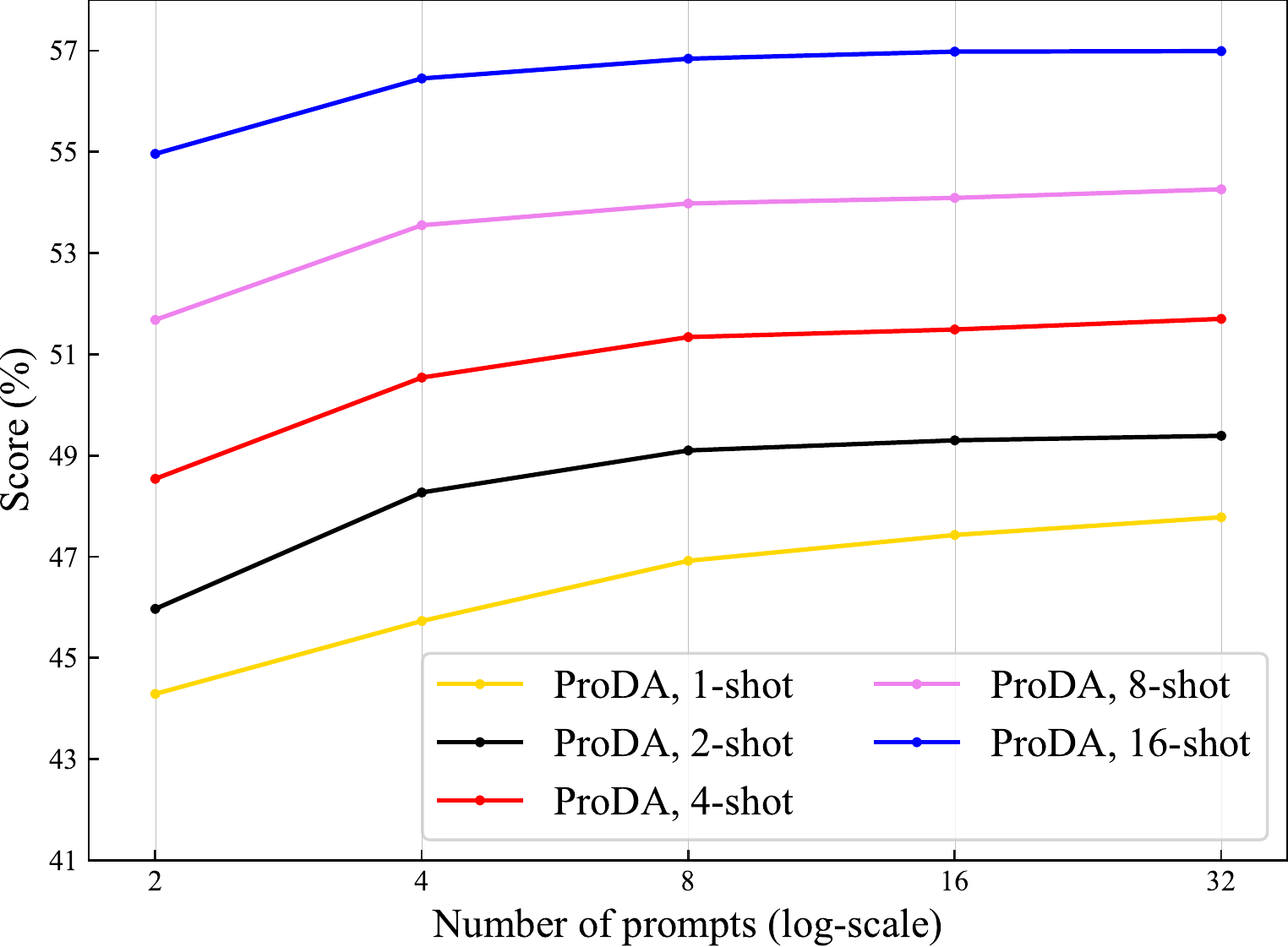}
   \caption{\textbf{Number of prompts.} We show the results of our ProDA on CIFAR-100 \cite{krizhevsky2009learning} with different shots.
   Larger prompt collection results in improvements. More prompts enable more precise estimation of the prompt distribution.} 
   \label{fig:n_prompt}
   \vspace{-0.5cm}
\end{figure}

\paragraph{Number of learnable prompts.} Fig. \ref{fig:n_prompt} shows the effect of the numbers of prompts on CIFAR-100 \cite{krizhevsky2009learning} recognition results. More prompts can improve the performance on the downstream task.
Increasing the number of prompts brings more diverse descriptions, enabling sufficiently representing visual variations.

\vspace{-0.3cm}

\paragraph{Sampling mini-batch prompts.} 
We sample mini-batch prompts in each training iteration to reduce the memory overhead instead of using all prompts. As shown in Table \ref{exp:minibatch}, the sampling strategy has similar results compared to using all prompts. However, using all prompts requires eight times the size of input texts, limited by the GPU memory size.

\section{Discussion and Conclusion}
\label{sec:conc}

This paper proposes a novel prompt learning method that learns the distribution of diverse prompts to address downstream visual recognition tasks with a pre-trained VLM.
Prompt learning is naturally suited to the problem of natural language, attracting significant attention recently. We believe it is also crucial to computer vision and could be a promising way to address vision tasks efficiently. 
The information of an images is not as abstract as language, which exacerbates the difficulty of learning a concept with limited visual supervision.
In contrast, language generated by humans has dense information and semantics. 
In this way, a few text descriptions are capable of providing considerable task-related content.
Our method demonstrates substantial improvement over the linear probe, which is a strong baseline of few-shot learning.
We hope our approach will inspire future work.

\paragraph{Limitation.} 
Prompt distribution learning proposed in this paper focuses on object/image recognition.
Computer vision has many other tasks, such as object detection, semantic segmentation, image style transfer, \textit{etc}. Our current methods cannot be applied to these tasks.
We believe that with dedicated modifications, our method can help some of them, which will be studied in future work.

\section*{Acknowledgement}
The research was partially supported by the National Natural Science
Foundation of China No. 61872329, and by MindSpore \cite{mindspore} which is a new deep learning computing framework.

{\small
\bibliographystyle{ieee_fullname}
\bibliography{proda}
}

\clearpage
\newpage
\appendix

\section{Proof of Proposition 1}
\begin{proposition}
Suppose that $\mathbf{w}_{1:C}=[\mathbf{w}^T_1,..,\mathbf{w}^T_C]^T\in\mathbb{R}^{dC}$ follows $\mathcal{N}(\mathbf{\mu}_{1:C}(\mathcal{P}^K), \mathbf{\Sigma}_{1:C}(\mathcal{P}^K))$. Let $\mathbf{\Sigma}_{ij}(\mathcal{P}^K)$ be the covariance matrix of $\mathbf{w}_i$ and $\mathbf{w}_j$, $\mathbf{\mu}_{i}(\mathcal{P}^K)$ be the mean of $\mathbf{w}_i$, and $\mathbf{A}_{i,j}=\mathbf{\Sigma}_{ii} + \mathbf{\Sigma}_{jj}-\mathbf{\Sigma}_{ij}-\mathbf{\Sigma}_{ji}$. Then it holds that
\begin{align}
    \mathcal{L}(\mathcal{P}^K) &\leq \mathop{\mathbb{E}}\limits_{\mathbf{x}_i,y_i} \left[ - \log \frac{e^{\mathbf{z}_i^T \mathbf{\mu}_{y_i}(\mathcal{P}^K)/\tau}}{\sum_c e^{ \mathbf{z}_i^T \mathbf{\mu}_c(\mathcal{P}^K)/\tau + \mathbf{z}^T_i \mathbf{A}_{c,y_i} \mathbf{z}_i / 2\tau^2} } \right]\\
    &\triangleq \mathcal{L}_{upper}(\mathcal{P}^K).
\label{equ:proof1}
\end{align}
\paragraph{Proof:} Given the Gaussian distribution assumption, we have:
\begin{align}
    \mathcal{L}(\mathcal{P}^K) &= \mathop{\mathbb{E}}\limits_{\mathbf{x}_i,y_i} \left[ - \log \mathop{\mathbb{E}}\limits_{\mathbf{w}_{1:C}} \frac{e^{\mathbf{z}_i^T \mathbf{w}_{y_i} / \tau}}{\sum_c e^{\mathbf{z}_i^T \mathbf{w}_c /\tau }} \right] \\
    \label{ieq2} & \leq \mathop{\mathbb{E}}\limits_{\mathbf{x}_i,y_i} \left[ \mathop{\mathbb{E}}\limits_{\mathbf{w}_{1:C}} \left[ - \log  \frac{e^{\mathbf{z}_i^T \mathbf{w}_{y_i} / \tau}}{\sum_c e^{\mathbf{z}_i^T \mathbf{w}_c /\tau }} \right]\right] \\
    & = \mathop{\mathbb{E}}\limits_{\mathbf{x}_i,y_i} \left[ \mathop{\mathbb{E}}\limits_{\mathbf{w}_{1:C}} \left[ \log \sum_c e^{\mathbf{z}_i^T (\mathbf{w}_c-\mathbf{w}_{y_i}) /\tau }  \right]\right] \\
    \label{ieq:1} & \leq \mathop{\mathbb{E}}\limits_{\mathbf{x}_i,y_i} \left[  \log \sum_c \mathop{\mathbb{E}}\limits_{\mathbf{w}_{1:C}} \left[ e^{\mathbf{z}_i^T (\mathbf{w}_c-\mathbf{w}_{y_i}) /\tau }  \right]\right] \\
    \label{eq:7}&= \mathop{\mathbb{E}}\limits_{\mathbf{x}_i,y_i} \left[  \log \sum_c e^{ \mathbf{z}_i^T (\mathbf{\mu}_c-\mathbf{\mu}_{y_i})/\tau + \mathbf{z}^T_i \mathbf{A}_{c,y_i} \mathbf{z}_i / 2\tau^2}
     \right] \\
     &= \mathop{\mathbb{E}}\limits_{\mathbf{x}_i,y_i} \left[ - \log \frac{e^{\mathbf{z}_i^T \mathbf{\mu}_{y_i}(\mathcal{P}^K)/\tau}}{\sum_c e^{ \mathbf{z}_i^T \mathbf{\mu}_c(\mathcal{P}^K)/\tau + \mathbf{z}^T_i \mathbf{A}_{c,y_i} \mathbf{z}_i / 2\tau^2} } \right],
\label{equ:proof2}
\end{align}
where Inequalities \ref{ieq2} and \ref{ieq:1} are from Jensen’s inequality ($\mathop{\mathbb{E}}(\log X) \leq \log \mathop{\mathbb{E}}(X)$ \cite{Jensen1906}). Besides, due to $\mathbf{z}_i^T (\mathbf{w}_c-\mathbf{w}_{y_i})$ being a Gaussian variable following $\mathcal{N}(\mathbf{z}_i^T (\mathbf{\mu}_c-\mathbf{\mu}_{y_i}), \mathbf{z}^T_i \mathbf{A}_{c,y_i} \mathbf{z}_i)$, the expectation in Equation \ref{eq:7} is
obtained by leveraging the moment-generating function:
\begin{equation}
    \mathop{\mathbb{E}}(e^{tX}) = \mathop{\mathbb{E}}(e^{t\mu + \frac{1}{2}\sigma^2 t^2}), X \sim \mathcal{N}(\mu, \sigma^2).
\end{equation}
\end{proposition}

\section{Datasets}
The details of the $12$ downstream datasets are shown in Tabel \ref{tab:dataset}. The accuracy metric of each dataset follows CLIP \cite{radford2021learning}.

\section{Baselines}
The regularization of Linear Probe CLIP is selected by the validation set on each dataset, following the hyperparameter sweep strategy in CLIP \cite{radford2021learning}. Note that the validation set, which is used to select task-specific hyperparameters, is only used in Linear Probe CLIP.

For CoOp \cite{zhou2021learning}, we use the SGD optimizer with learning rate of $0.001$ and the batch size of $20$. The learning rate
has a cosine decay schedule. The number of training epochs is $100$. The prompt length is $16$.
Our implementation of Prompt Tuning has slightly better results than those reported in \cite{zhou2021learning}.

\section{Results}

Table \ref{tab:res} shows the detailed results of various methods with the same pre-trained CLIP model (RN50) on the 12 datasets.

\begin{table*}[tbp]
    \begin{center}
        \begin{tabular}{lrrrr}
        \hline
        Dataset  & Classes & Train Size & Test Size & Accuracy metric\\
        \hline
        ImageNet \cite{deng2009imagenet} & 1000 & 1,281,167 & 50,000 & accuracy \\
        CIFAR-10 \cite{krizhevsky2009learning} & 10 & 50,000 & 10,000 & accuracy\\
        CIFAR-100 \cite{krizhevsky2009learning}& 100 & 50,000 & 10,000 & accuracy\\
        STL-10 \cite{coates2011an} & 10 & 1,000 & 8,000 & accuracy \\
        Food-101 \cite{bossard2014food} & 101 & 75,750 & 25,250 & accuracy\\
        Stanford Cars \cite{krause20133d} & 196 & 8,144 & 8,041 & accuracy\\
        FGVC Aircraft \cite{maji13fine-grained} & 100 & 6,667 & 3,333 & mean per-class\\
        Oxford-IIIT Pets \cite{parkhi2012cats} & 37 & 3,680 & 3,669 & mean per-class\\
        Caltech-101 \cite{li2017learning} & 102 & 3,060 & 6,086 & mean per-class\\
        Oxford 102 Flowers \cite{nilsback2008automated} & 102 & 2,040 & 6,149 & mean per-class\\
        EuroSAT \cite{helber2019eurosat} & 10 & 10,000 & 5,000 & accuracy \\
        Describable Textures (DTD) \cite{cimpoi2014describing} & 47 & 3,760 & 1,880 & accuracy\\
        \hline
        \end{tabular}
    \end{center}
    \caption{Datasets in our experiments.}
    \label{tab:dataset}
    \label{datasets}
\end{table*}

\begin{table*}[htbp]
    \centering
    \footnotesize
    \begin{tabular}{cccccccccccccc}
        Method & \# Shot & 
        \rotatebox{90}{ImageNet} & 
        \rotatebox{90}{CIFAR-10} & 
        \rotatebox{90}{CIFAR-100} & 
         \rotatebox{90}{STL-10} &
         \rotatebox{90}{Food-101} &
         \rotatebox{90}{Stanford Cars} &
         \rotatebox{90}{FGVC Aircraft} &
         \rotatebox{90}{Oxford Pets} &
         \rotatebox{90}{Caltech-101} &
         \rotatebox{90}{Oxford Flowers} &
         \rotatebox{90}{EuroSAT} &
         \rotatebox{90}{DTD} \\
         \hline
       Zero-Shot CLIP & 0  & 59.8 & 71.6 & 40.6 & 94.4 & 80.6 & 54.3 & 17.0 & 85.5 & 84.5 & 65.5 & 41.8 & 41.2 \\
       \hline
                   & 1  & 22.1 & 44.3 & 18.2 & 80.6 & 31.4 & 24.3 & 13.0 & 30.2 & 56.7 & 56.1 & 46.8 & 29.8 \\
                   & 2  & 31.9 & 53.5 & 26.6 & 86.9 & 45.1 & 36.8 & 17.9 & 40.5 & 68.7 & 74.9 & 55.5 & 41.4 \\
Linear Probe CLIP  & 4  & 41.4 & 62.2 & 35.6 & 92.2 & 56.8 & 49.5 & 23.9 & 56.4 & 79.0 & 86.3 & 65.7 & 51.9 \\
                   & 8  & 49.4 & 70.1 & 43.6 & 94.3 & 67.1 & 61.2 & 29.4 & 67.5 & 84.3 & 91.5 & 75.1 & 59.0 \\
                   & 16 & 55.9 & 73.8 & 50.5 & 95.0 & 73.7 & 70.0 & 36.0 & 75.1 & 87.3 & 95.6 & 80.7 & 64.3 \\
           \hline
                   & 1  & 53.4 & 71.8 & 41.3 & 94.1 & 77.5 & 54.0 & 17.7 & 86.5 & 83.6 & 70.0 & 47.1 & 44.1 \\
                   & 2  & 55.7 & 73.2 & 42.4 & 94.3 & 76.4 & 57.0 & 19.9 & 86.7 & 84.2 & 77.8 & 60.4 & 48.4 \\
CoOp      & 4  & 57.9 & 75.4 & 45.7 & 94.9 & 77.0 & 61.4 & 22.7 & 87.2 & 85.5 & 84.8 & 69.4 & 53.4 \\
                   & 8  & 60.5 & 76.7 & 49.7 & 95.3 & 77.8 & 65.5 & 26.3 & 87.8 & 87.4 & 89.9 & 76.5 & 58.7 \\
                   & 16 & 62.3 & 78.4 & 53.3 & 95.6 & 79.3 & 70.5 & 30.1 & 88.4 & 89.6 & 93.4 & 81.4 & 65.0 \\
           \hline
                   & 1  & 61.8 & 74.6 & 47.8 & 95.1 & 80.8 & 60.1 & 22.2 & 88.2 & 86.7 & 77.5 & 58.5 & 50.9 \\
                   & 2  & 62.3 & 76.4 & 49.4 & 95.3 & 80.6 & 63.7 & 24.8 & 88.4 & 87.1 & 85.2 & 69.0 & 56.2 \\
ProDA (ours)       & 4  & 63.6 & 78.3 & 51.7 & 95.7 & 80.8 & 67.9 & 27.5 & 89.0 & 88.7 & 90.6 & 75.3 & 60.0 \\
                   & 8  & 64.7 & 79.6 & 54.3 & 96.1 & 81.7 & 72.1 & 31.5 & 89.4 & 89.8 & 93.6 & 80.1 & 65.0 \\
                   & 16 & 65.3 & 80.9 & 57.0 & 96.3 & 82.4 & 75.5 & 36.6 & 90.0 & 91.3 & 95.5 & 84.3 & 70.1 \\
           \hline
       
    \end{tabular}
    \caption{Detailed performance (\%) of various methods on the $12$ downstream datasets. ``\# Shot'' denotes the number of training samples per class.}
    \label{tab:res}
\end{table*}

\end{document}